\documentclass[conference]{IEEEtran}
\IEEEoverridecommandlockouts

\usepackage{cite}
\usepackage{amsmath,amssymb,amsfonts}
\usepackage{graphicx}
\usepackage{textcomp}
\usepackage{xcolor}
\usepackage{pgfplots}
\usepackage{fancyhdr}
\usepackage{algorithm}
\usepackage{algorithmicx}
\usepackage{algpseudocode}
\usepackage{pifont} 

\def\BibTeX{{\rm B\kern-.05em{\sc i\kern-.025em b}\kern-.08em
    T\kern-.1667em\lower.7ex\hbox{E}\kern-.125emX}}
\begin{document}

\title{Intelligent Task Offloading in VANETs: A Hybrid AI-Driven Approach for Low-Latency and Energy Efficiency}
\author{
    \IEEEauthorblockN{Tariq Qayyum\IEEEauthorrefmark{1},
    Asadullah Tariq\IEEEauthorrefmark{1},
    Muhammad Ali\IEEEauthorrefmark{2}, Mohamed Adel Serhani\IEEEauthorrefmark{3}, Zouheir Trabelsi\IEEEauthorrefmark{1}, \\Maite López-Sánchez\IEEEauthorrefmark{4}}

   \\
  \IEEEauthorblockA{\IEEEauthorrefmark{1}College of Information Technology, United Arab Emirates University, Al Ain, Abu Dhabi, UAE\\
    }
    \IEEEauthorblockA{\IEEEauthorrefmark{2}Department of Computer Science, SCIT, Beaconhouse National University, Lahore Pakistan\\
    }
    
    \IEEEauthorblockA{\IEEEauthorrefmark{3}College of Computing and Informatics, University of Sharjah, Sharjah, UAE\\
    }

    \IEEEauthorblockA{\IEEEauthorrefmark{4}Department of Mathematics, University of Barcelona, Gran Via de les Corts Catalanes, Barcelona\\
    }

}

\maketitle

\fancypagestyle{firstpageheader}{
    \fancyhf{} 
    \renewcommand{\headrulewidth}{0pt} 
    \fancyhead[C]{\small Accepted for Presentation at International Wireless Communications \& Mobile Computing Conference (IWCMC) 2025, Abu Dhabi, UAE.} 
}
\thispagestyle{firstpageheader}
\begin{abstract}
Vehicular Ad-hoc Networks (VANETs) are integral to intelligent transportation systems, enabling vehicles to offload computational tasks to nearby roadside units (RSUs) and mobile edge computing (MEC) servers for real-time processing. However, the highly dynamic nature of VANETs introduces challenges, such as unpredictable network conditions, high latency, energy inefficiency, and task failure. This research addresses these issues by proposing a hybrid AI framework that integrates supervised learning, reinforcement learning, and Particle Swarm Optimization (PSO) for intelligent task offloading and resource allocation. The framework leverages supervised models for predicting optimal offloading strategies, reinforcement learning for adaptive decision-making, and PSO for optimizing latency and energy consumption. Extensive simulations demonstrate that the proposed framework achieves significant reductions in latency and energy usage while improving task success rates and network throughput. By offering an efficient, and scalable solution, this framework sets the foundation for enhancing real-time applications in dynamic vehicular environments.
\end{abstract}


\section{Introduction}

Vehicular Ad-hoc Networks (VANETs) play a crucial role in modern intelligent transportation systems by enabling vehicle-to-vehicle (V2V) and vehicle-to-infrastructure (V2I) communication. These networks enhance road safety, traffic efficiency, and autonomous driving by facilitating real-time data exchange. However, the increasing reliance on advanced sensors, cameras, and computational modules in vehicles has led to a surge in data processing demands~\cite{1}. Many applications, including collision avoidance, traffic monitoring, and in-vehicle infotainment, generate significant computational workloads that exceed the processing capabilities of individual vehicles. Computation offloading, wherein tasks are dynamically allocated to Roadside Units (RSUs), Mobile Edge Computing (MEC) servers, or cloud infrastructures, emerges as a viable solution to mitigate computational constraints and improve service responsiveness~\cite{2}.

Despite its potential, offloading decisions in VANETs present significant challenges due to their highly dynamic and decentralized nature. Rapid topology changes, intermittent connectivity, variable computational loads, and stringent latency requirements complicate the decision-making process. Traditional rule-based or heuristic offloading strategies often struggle to adapt to real-time traffic variations, resulting in increased latency, suboptimal resource utilization, and inefficient energy consumption. Therefore, an intelligent and adaptive mechanism is required to balance local execution and offloading, considering factors such as network conditions, vehicle mobility, and available computational resources~\cite{3}.

Artificial Intelligence (AI) is a powerful tool for managing task offloading in VANETs, leveraging data to predict network conditions and make adaptive decisions. Hybrid frameworks combining supervised learning, deep learning, and reinforcement learning enhance offloading efficiency. Supervised models like Random Forest predict feasibility, while CNNs and LSTMs analyze traffic patterns. RL dynamically learns optimal policies, and heuristic techniques such as PSO optimize resource allocation to reduce latency and energy consumption~\cite{4}, ~\cite{5}.

This research introduces a novel hybrid AI framework that addresses the limitations of traditional offloading approaches. By integrating predictive analytics, real-time adaptability, and heuristic optimization, the proposed framework enhances computational efficiency, reduces latency, and balances energy consumption. Furthermore, the incorporation of federated learning principles allows AI models to be collaboratively trained across multiple vehicles, improving decision-making while preserving data privacy and reducing communication overhead.

\textbf{Contributions:} The key contributions of this research are: (1) the development of a predictive offloading model that dynamically determines optimal execution strategies based on real-time traffic conditions and resource availability, (2) an RL-based adaptive offloading mechanism that continuously learns optimal policies through real-world traffic simulations, and (3) the application of bio-inspired optimization techniques to enhance resource allocation efficiency. Extensive simulations demonstrate that the proposed framework significantly improves latency reduction, energy efficiency, and network throughput compared to conventional offloading strategies.

\begin{figure*}[h!]
    \centering
    \includegraphics[width=0.87\textwidth]{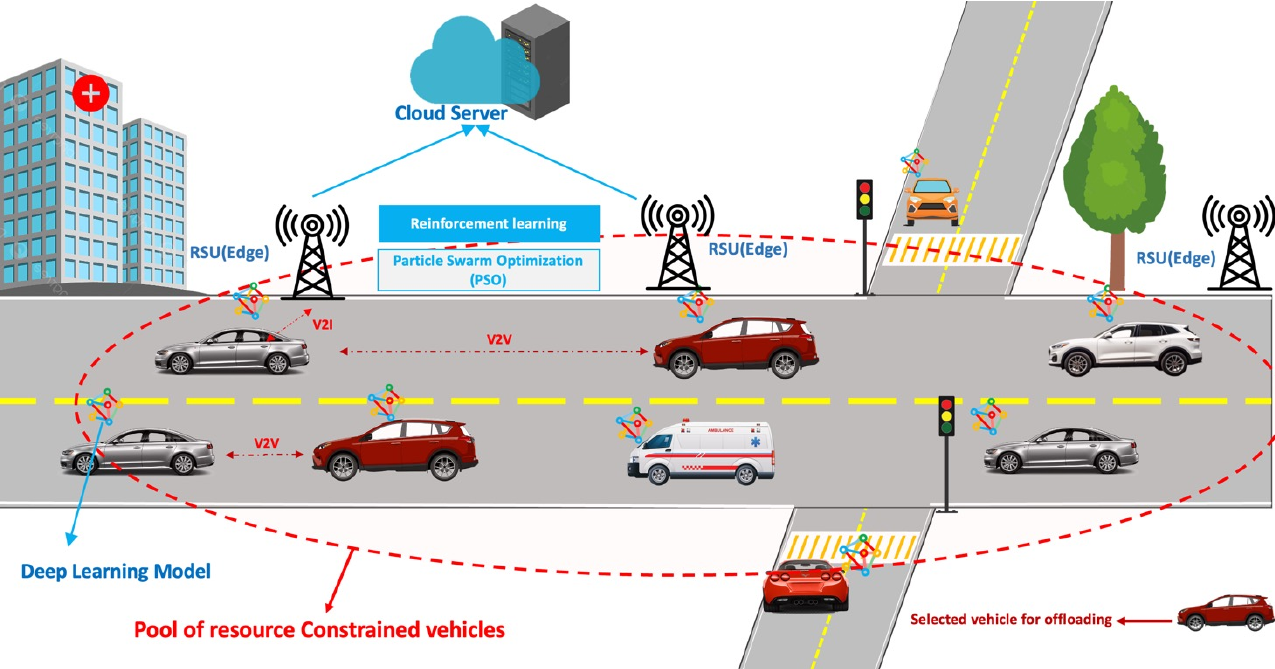} 
    \caption{Overview of the proposed system for intelligent task offloading in VANETs. The system integrates reinforcement learning and Particle Swarm Optimization (PSO) for optimal task distribution between vehicles and RSUs, ensuring low latency and energy efficiency.}
    \label{fig:system_architecture}
\end{figure*}

By addressing the inherent challenges of task offloading in dynamic vehicular environments, this research advances AI-driven decision-making for scalable and intelligent VANET architectures. The proposed framework provides a roadmap for the integration of AI-based solutions in next-generation vehicular networks, paving the way for more efficient and autonomous transportation systems. Figure~\ref{fig:system_architecture} shows the proposed system architecture that includes AI based task offloading leveraging PSO and RL. 

\section{Related Work}
In this section, we present a state-of-the-art review of AI-based approaches for task offloading in VANETs.
Firdose et al.~\cite{6} proposed a comprehensive survey addressing task offloading challenges in Edge and Cloud Computing. The study highlights the limitations of centralized Cloud systems for delay-sensitive applications and emphasizes the potential of Edge Computing to mitigate these issues. 

Yen et al.~\cite{7} focuses on a game-theoretic approach and adaptive type selection for resource sharing, which may not effectively handle the high variability in VANET environments. Unlike our proposed hybrid AI framework, which integrates supervised learning, reinforcement learning, and PSO for intelligent task offloading, the existing model lacks a comprehensive AI-driven optimization mechanism, leading to potential inefficiencies in latency reduction, energy optimization, and adaptive decision-making in dynamic vehicular scenarios.

Han et al.~\cite{8} proposed an intelligent computational offloading scheme, CODIA, to address the NP-hard problem of dependent task offloading in dynamic IoT environments. CODIA employs a prioritized subtask scheduling strategy and an Actor-Critic-based reinforcement learning algorithm for efficient decision-making. It transforms task dependency into device state transitions, ensuring low latency and energy efficiency. 

Guanjin et al.~\cite{9} proposed the Deep Meta Reinforcement Learning-based Offloading (DMRO) framework, which integrates DRL and meta-learning for efficient IoT task offloading in dynamic MEC environments. DMRO achieves faster adaptation, strong portability, and 17.6\% better performance over traditional DRL algorithms.

Ying et al.~\cite{10} propose the Efficient Multi-Vehicle Task Offloading (EMT) algorithm for 6G networks using Cybertwin-enabled MEC. EMT minimizes system costs and queue lengths, balancing performance with dynamic green energy utilization.

Kai et al.~\cite{11} introduces a task offloading framework in heterogeneous vehicular networks, leveraging V2X communication technologies and federated Q-learning to minimize resource costs and offloading failure probabilities, ensuring optimal system performance. Moreover, authors in \cite{12} propose an efficient SDN-assisted forwarding strategy within NDN-based IoV, designed to support vehicular mobility while leveraging cellular networks to ensure low-latency control message delivery

Trabelsi et al.~\cite{13} propose a fuzzy-based V2V offloading method for low-latency, priority-aware IoV tasks.
Their approach optimizes task scheduling, ensuring reduced system delay and efficient resource utilization.

Ai et al.~\cite{14} proposed a hybrid optimization framework using reinforcement learning and deep reinforcement learning to minimize power consumption in MEC systems while ensuring low-latency task offloading and resource allocation.

\section{System Model}
Vehicular Ad-hoc Networks (VANETs) have emerged as a cornerstone of modern intelligent transportation systems, enabling vehicles to communicate with each other and with roadside infrastructure to enhance road safety, traffic efficiency, and autonomous driving capabilities. As vehicles are increasingly equipped with advanced sensors, cameras, and computational modules, the demand for real-time processing of massive amounts of data has surged. Applications such as autonomous driving, collision avoidance, traffic monitoring, and in-vehicle infotainment generate extensive computational workloads that often exceed the limited processing capabilities of individual vehicles. Computation offloading, wherein tasks are dynamically allocated to nearby Roadside Units (RSUs), Mobile Edge Computing (MEC) servers\cite{15}, or cloud-based infrastructures, presents a promising solution to alleviate the computational burden on vehicles and enhance service responsiveness.

In the proposed framework, we consider a vehicular network consisting of a set of vehicles \( V = \{ v_1, v_2, \dots, v_n \} \) and a set of roadside units (RSUs) \( R = \{ R_1, R_2, \dots, R_m \} \). Each vehicle generates computational tasks that can either be executed locally or offloaded to an RSU, MEC server, or cloud for processing. The decision to offload is influenced by factors such as vehicle speed, channel conditions, computational load, and latency requirements. Vehicles dynamically interact with RSUs, forming a highly dynamic and distributed computing system where intelligent task allocation is crucial for optimizing performance and reducing communication delays.

The total task workload generated by a vehicle \( v_i \) is represented as \( W_i \), which consists of multiple tasks characterized by their computation intensity \( C_i \) (in CPU cycles per bit) and data size \( D_i \) (in bits). The processing time for a task executed locally is given by:
\begin{equation}
T_{local} = \frac{C_i}{f_i},
\end{equation}
where \( f_i \) is the computational capability of the vehicle \( v_i \) in CPU cycles per second. 

For offloading, the transmission time to an RSU \( R_j \) is given by:
\begin{equation}
T_{tx} = \frac{D_i}{R_{ij}},
\end{equation}
where \( R_{ij} \) is the transmission rate between \( v_i \) and \( R_j \). The total time for offloading and execution at an RSU is:
\begin{equation}
T_{offload} = T_{tx} + \frac{C_i}{f_j},
\end{equation}
where \( f_j \) is the computational capability of the RSU.

The energy consumption for local execution is:
\begin{equation}
E_{local} = \kappa C_i f_i^2,
\end{equation}
where \( \kappa \) is a hardware-dependent energy coefficient. The energy consumption for offloading is:
\begin{equation}
E_{tx} = P_{tx} T_{tx},
\end{equation}
where \( P_{tx} \) is the transmission power of the vehicle.

The optimization objective is to minimize the total execution latency and energy consumption:
\begin{equation}
\min \sum_{i=1}^{n} (T_{offload} + \lambda E_{tx}),
\end{equation}
where \( \lambda \) is a weight factor balancing latency and energy consumption.

A reinforcement learning-based policy \( \pi \) is introduced to make intelligent offloading decisions. The expected reward function is:
\begin{equation}
R = \mathbb{E} \left[ - (T_{offload} + \lambda E_{tx}) \right],
\end{equation}
where the RL agent optimizes the policy \( \pi \) to maximize this reward.

The wireless communication model assumes a channel gain \( h_{ij} \) with path loss exponent \( \alpha \):
\begin{equation}
R_{ij} = B \log_2 \left( 1 + \frac{P_{tx} h_{ij}}{N_0} \right),
\end{equation}
where \( B \) is the bandwidth and \( N_0 \) is the noise power spectral density.

Table \ref{table:notations} summarizes the main notations used in the system model.

\begin{table}[h]
\centering
\caption{List of Notations}
\begin{tabular}{|c|c|}
\hline
Symbol & Description \\
\hline
$V$ & Set of vehicles \\
$R$ & Set of RSUs \\
$C_i$ & Computation intensity (CPU cycles per bit) \\
$D_i$ & Data size (bits) \\
$f_i$ & Computational capability of vehicle $v_i$ \\
$f_j$ & Computational capability of RSU $R_j$ \\
$T_{local}$ & Local execution time \\
$T_{offload}$ & Offloading execution time \\
$R_{ij}$ & Transmission rate between vehicle $v_i$ and RSU $R_j$ \\
$E_{local}$ & Energy consumption for local execution \\
$E_{tx}$ & Energy consumption for task transmission \\
$\lambda$ & Weight factor balancing latency and energy \\
$P_{tx}$ & Transmission power of vehicle \\
$h_{ij}$ & Wireless channel gain \\
$B$ & Bandwidth \\
\hline
\end{tabular}
\label{table:notations}
\end{table}

Vehicular Ad-hoc Networks (VANETs) have emerged as a cornerstone of modern intelligent transportation systems, enabling vehicles to communicate with each other and with roadside infrastructure to enhance road safety, traffic efficiency, and autonomous driving capabilities. As vehicles are increasingly equipped with advanced sensors, cameras, and computational modules, the demand for real-time processing of massive amounts of data has surged. Applications such as autonomous driving, collision avoidance, traffic monitoring, and in-vehicle infotainment generate extensive computational workloads that often exceed the limited processing capabilities of individual vehicles. Computation offloading, wherein tasks are dynamically allocated to nearby Roadside Units (RSUs), Mobile Edge Computing (MEC) servers, or cloud-based infrastructures, presents a promising solution to alleviate the computational burden on vehicles and enhance service responsiveness.

In the proposed framework, we consider a vehicular network consisting of a set of vehicles \( V = \{ v_1, v_2, \dots, v_n \} \) and a set of roadside units (RSUs) \( R = \{ R_1, R_2, \dots, R_m \} \). Each vehicle generates computational tasks that can either be executed locally or offloaded to an RSU, MEC server, or cloud for processing. The decision to offload is influenced by factors such as vehicle speed, channel conditions, computational load, and latency requirements. Vehicles dynamically interact with RSUs, forming a highly dynamic and distributed computing system where intelligent task allocation is crucial for optimizing performance and reducing communication delays.

The total task workload generated by a vehicle \( v_i \) is represented as \( W_i \), which consists of multiple tasks characterized by their computation intensity \( C_i \) (in CPU cycles per bit) and data size \( D_i \) (in bits). The processing time for a task executed locally is given by:
\begin{equation}
T_{local} = \frac{C_i}{f_i},
\end{equation}
where \( f_i \) is the computational capability of the vehicle \( v_i \) in CPU cycles per second. 

For offloading, the transmission time to an RSU \( R_j \) is given by:
\begin{equation}
T_{tx} = \frac{D_i}{R_{ij}},
\end{equation}
where \( R_{ij} \) is the transmission rate between \( v_i \) and \( R_j \). The total time for offloading and execution at an RSU is:
\begin{equation}
T_{offload} = T_{tx} + \frac{C_i}{f_j},
\end{equation}
where \( f_j \) is the computational capability of the RSU.

The energy consumption for local execution is:
\begin{equation}
E_{local} = \kappa C_i f_i^2,
\end{equation}
where \( \kappa \) is a hardware-dependent energy coefficient. The energy consumption for offloading is:
\begin{equation}
E_{tx} = P_{tx} T_{tx},
\end{equation}
where \( P_{tx} \) is the transmission power of the vehicle.

The optimization objective is to minimize the total execution latency and energy consumption:
\begin{equation}
\min \sum_{i=1}^{n} (T_{offload} + \lambda E_{tx}),
\end{equation}
where \( \lambda \) is a weight factor balancing latency and energy consumption.

A reinforcement learning-based policy \( \pi \) is introduced to make intelligent offloading decisions. The expected reward function is:
\begin{equation}
R = \mathbb{E} \left[ - (T_{offload} + \lambda E_{tx}) \right],
\end{equation}
where the RL agent optimizes the policy \( \pi \) to maximize this reward.

The wireless communication model assumes a channel gain \( h_{ij} \) with path loss exponent \( \alpha \):
\begin{equation}
R_{ij} = B \log_2 \left( 1 + \frac{P_{tx} h_{ij}}{N_0} \right),
\end{equation}
where \( B \) is the bandwidth and \( N_0 \) is the noise power spectral density.

Table \ref{table:notations} summarizes the main notations used in the system model. By integrating learning-based optimization and reinforcement learning policies, the proposed framework enables vehicles to make efficient, low-latency, and energy-aware offloading decisions in dynamic vehicular environments.

\begin{algorithm}
\caption{Reinforcement Learning-Based Offloading}
\begin{algorithmic}[1]
\Require Initial Q-function \( Q(s, a) \), policy \( \pi \), learning rate \( \alpha \), discount factor \( \gamma \), exploration probability \( \epsilon \)
\Ensure Optimized policy \( \pi^* \)
\State Initialize the RL agent with policy \( \pi \) and Q-function \( Q(s, a) \)
\State Set the environment state \( s \) with initial network parameters
\For{each episode}
    \State Reset the environment and observe initial state \( s \)
    \For{each time step}
        \State Select action \( a \) using \( \epsilon \)-greedy policy
        \State Execute action \( a \) and observe reward \( r \) and next state \( s' \)
        \State Update Q-function:
        \begin{equation}
        Q(s, a) \leftarrow Q(s, a) + \alpha \left[ r + \gamma \max_a Q(s', a) - Q(s, a) \right]
        \end{equation}
        \State Update policy \( \pi \) based on \( Q(s, a) \)
        \State Set \( s \leftarrow s' \)
    \EndFor
\EndFor
\end{algorithmic}
\end{algorithm}

\textbf{Reinforcement Learning-Based Offloading: }This algorithm dynamically determines the optimal offloading strategy for vehicles using reinforcement learning. The RL agent is initialized with a policy \( \pi \) and Q-function \( Q(s, a) \), which estimates the expected reward of taking an action \( a \) in a given state \( s \). The process iteratively updates the Q-function using observed rewards and state transitions based on the Bellman equation. This enables vehicles to adapt their offloading decisions to dynamic vehicular environments.

\begin{algorithm}
\caption{Optimization-Based Resource Allocation}
\begin{algorithmic}[1]
\Require Task parameters \( T_{offload}, E_{tx} \), initial particle population, velocity and position settings, iteration limit
\Ensure Optimized resource allocation strategy
\State Initialize population for Particle Swarm Optimization (PSO)
\State Set particle velocities and positions randomly
\For{each iteration}
    \For{each particle}
        \State Evaluate fitness function:
        \begin{equation}
        F = T_{offload} + \lambda E_{tx}
        \end{equation}
        \State Update particle velocity and position based on best solutions
    \EndFor
    \State Update global and local best solutions
\EndFor
\State Return optimized resource allocation strategy
\end{algorithmic}
\end{algorithm}

\textbf{Optimization-Based Resource Allocation: }This algorithm uses Particle Swarm Optimization (PSO) to minimize the objective function \( F = T_{offload} + \lambda E_{tx} \). Each particle represents a candidate solution, and their positions and velocities are iteratively updated based on their fitness values, as well as global and local best solutions. The process continues until convergence or the maximum iteration limit is reached, providing an optimized resource allocation strategy for VANETs.

\section{Experimental Setup}

To evaluate the performance of the proposed hybrid AI-based offloading framework, we conducted extensive simulations using a custom-built VANET simulation environment. The key parameters and configurations used in the experiments are summarized below:

The simulations were performed using SUMO (Simulation of Urban MObility) for vehicular mobility modeling and Omnet++ for network communication modeling. Realistic traffic scenarios, including urban and highway environments, were simulated to reflect varying vehicular densities and mobility patterns. The VANET environment consisted of 200 vehicles and 15 RSUs distributed across the simulation area. Each vehicle was equipped with a single wireless interface operating in the 5.9 GHz band, with a transmission range of 300 meters.

Each vehicle generated computational tasks at random intervals with task sizes ranging from 1 MB to 10 MB. The computational intensity of tasks varied between 500 and 1000 CPU cycles per bit. The hybrid AI framework included supervised learning for predictive task offloading, reinforcement learning for adaptive decision-making, and Particle Swarm Optimization (PSO) for resource allocation optimization. Federated learning was employed for collaborative model training across vehicles without sharing raw data.

\section{Proposed Plots and Analysis}

\subsection{Latency and Energy Comparison}
\begin{figure}[h]
    \centering
    \includegraphics[width=0.45\textwidth, height=0.25\textwidth]{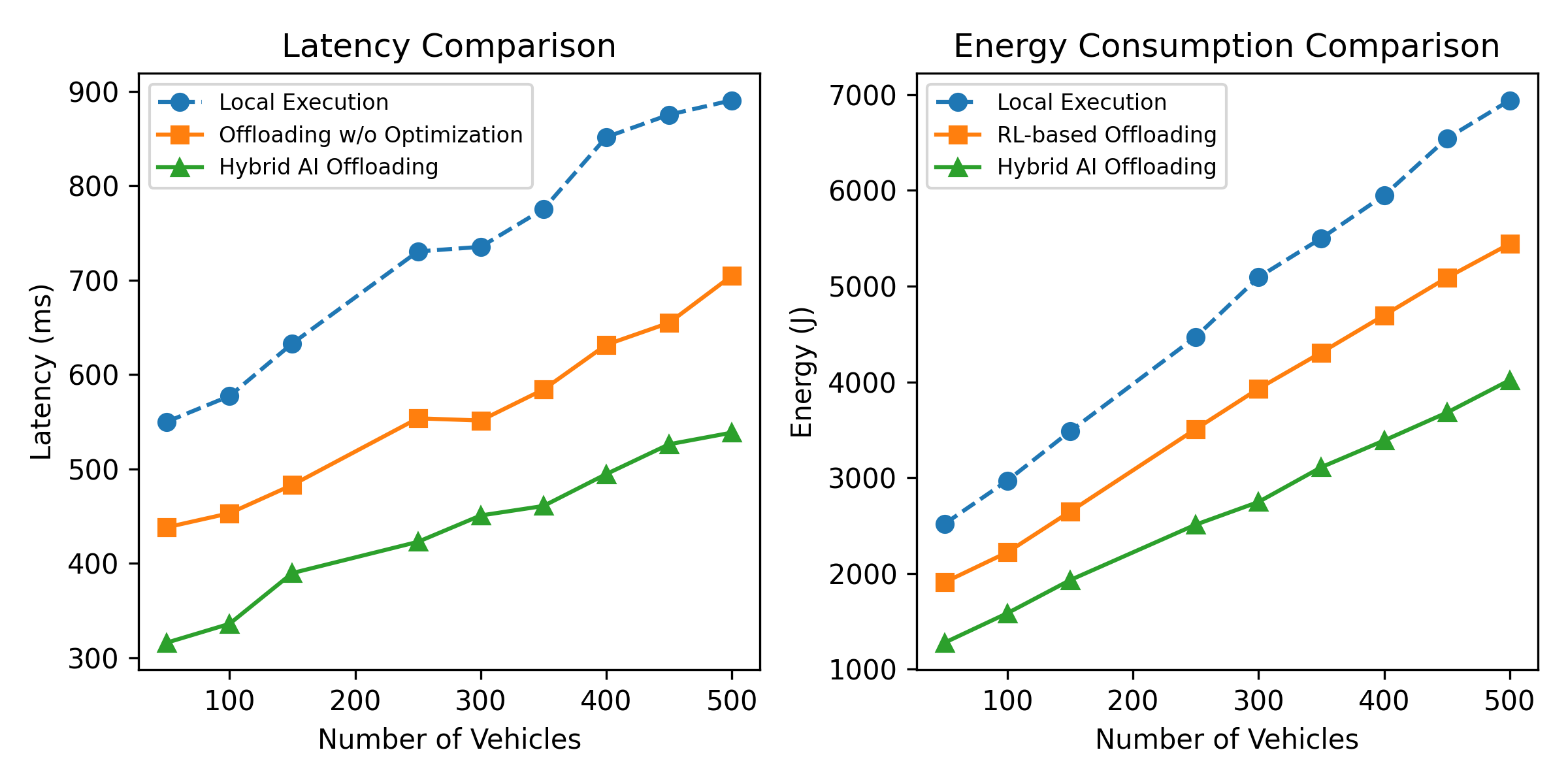}
    \caption{Latency and Energy Consumption Comparison}
    \label{fig:latency_energy}
\end{figure}
The proposed hybrid AI framework demonstrates significantly reduced latency compared to local execution and offloading without optimization, as shown in Figure~\ref{fig:latency_energy}. By dynamically learning network conditions and optimizing task allocation, our framework minimizes both latency and energy consumption, showcasing superior efficiency over traditional methods.

\subsection{Task Offloading Ratio and Throughput}
\begin{figure}[h]
    \centering
    \includegraphics[width=0.45\textwidth, height=0.25\textwidth]{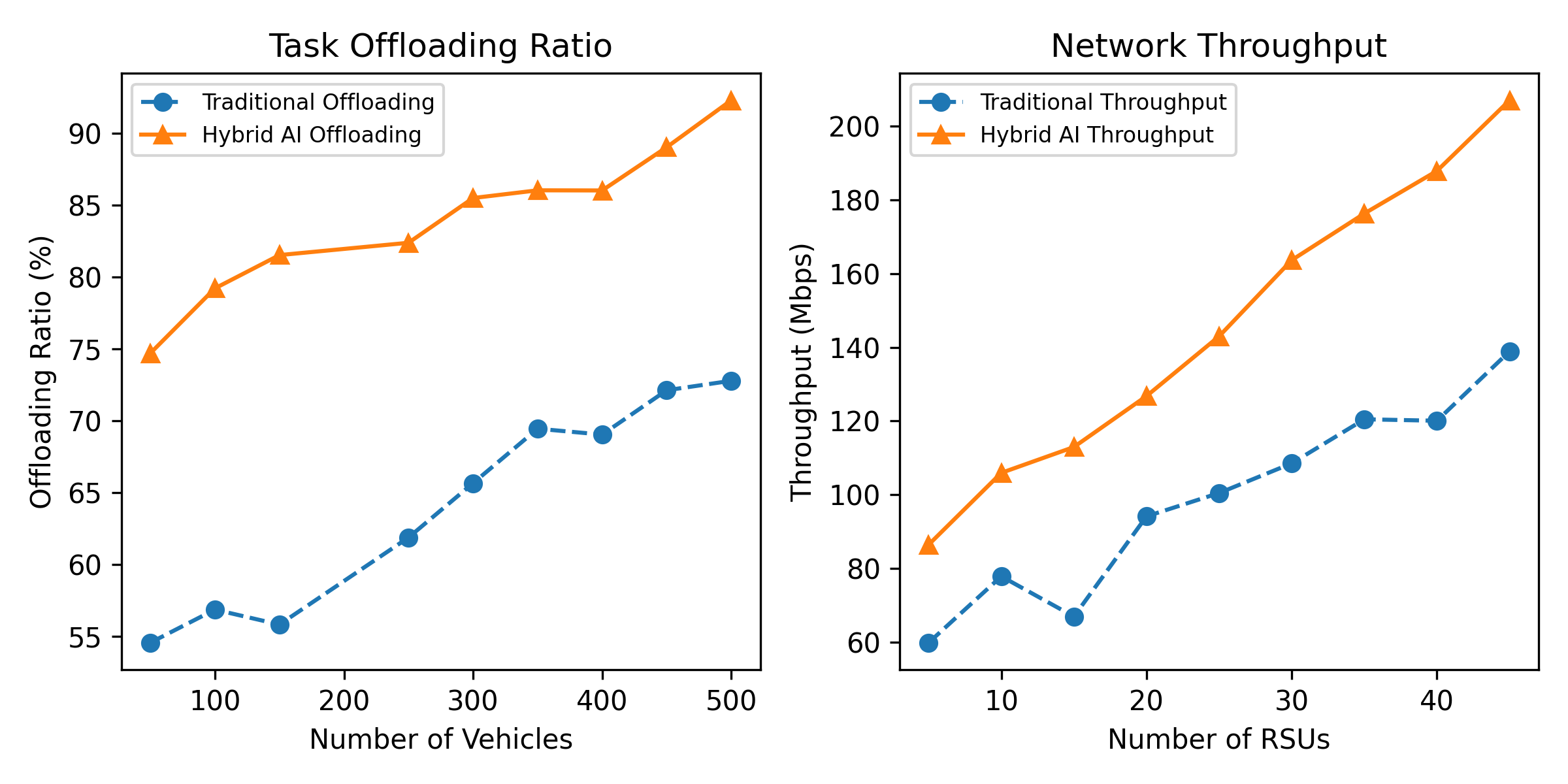}
    \caption{Task Offloading Ratio and Throughput}
    \label{fig:offloading_throughput}
\end{figure}
Figure~\ref{fig:offloading_throughput} highlights the adaptability of our framework, achieving higher task offloading ratios and network throughput. The intelligent task allocation ensures effective utilization of resources, avoiding bottlenecks and improving overall performance.

\subsection{Failure Rate and Channel Utilization}
\begin{figure}[h]
    \centering
    \includegraphics[width=0.45\textwidth, height=0.25\textwidth]{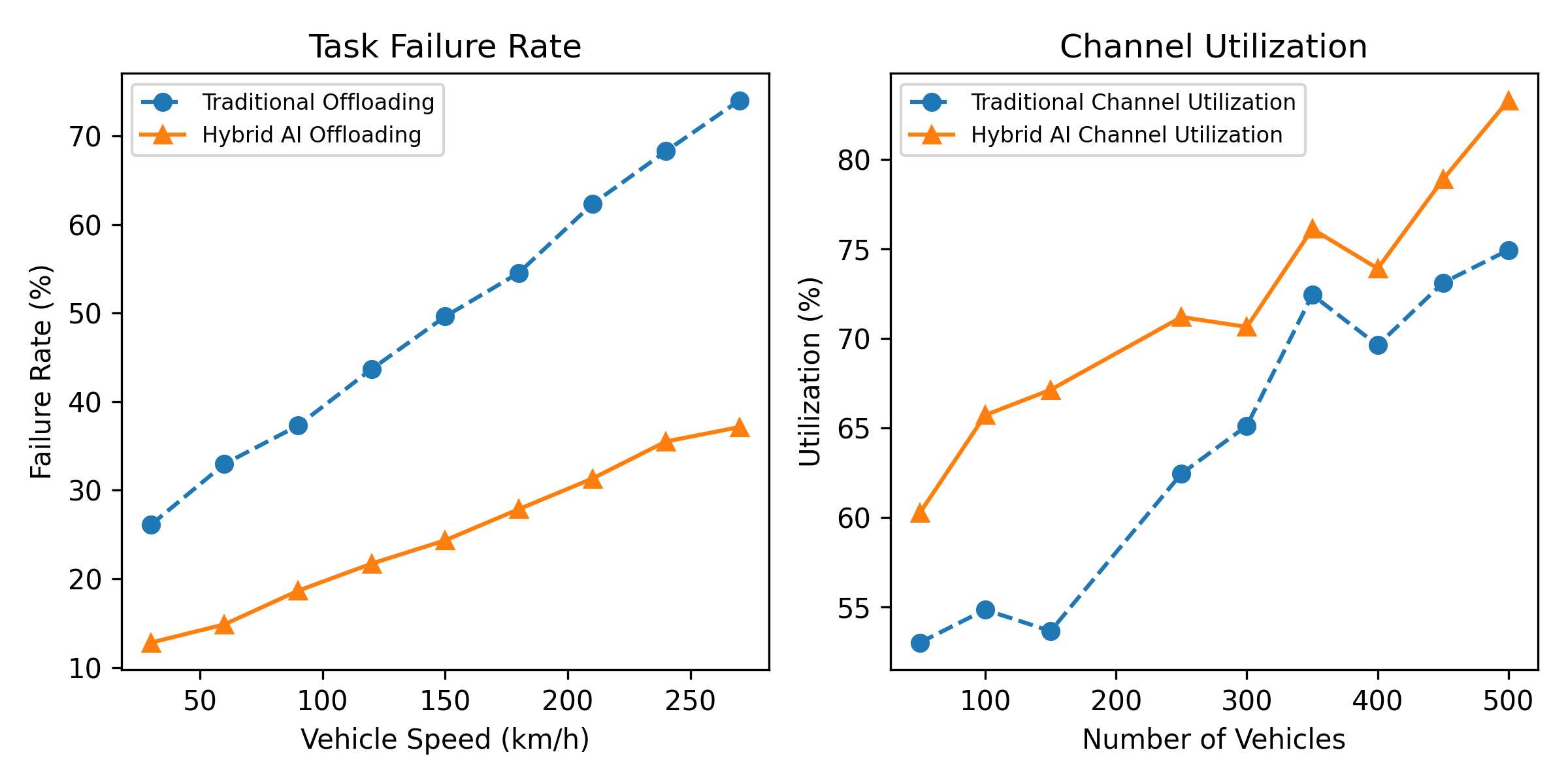}
    \caption{Task Failure Rate and Channel Utilization}
    \label{fig:failure_channel}
\end{figure}
Figure~\ref{fig:failure_channel} shows that our framework significantly reduces the task failure rate and improves channel utilization. These improvements result from the predictive capabilities of our system, which adapt to real-time network conditions and avoid overloading resources.

\subsection{Reward Convergence in Reinforcement Learning}
\begin{figure}[h]
    \centering
    \includegraphics[width=0.45\textwidth, height=0.25\textwidth]{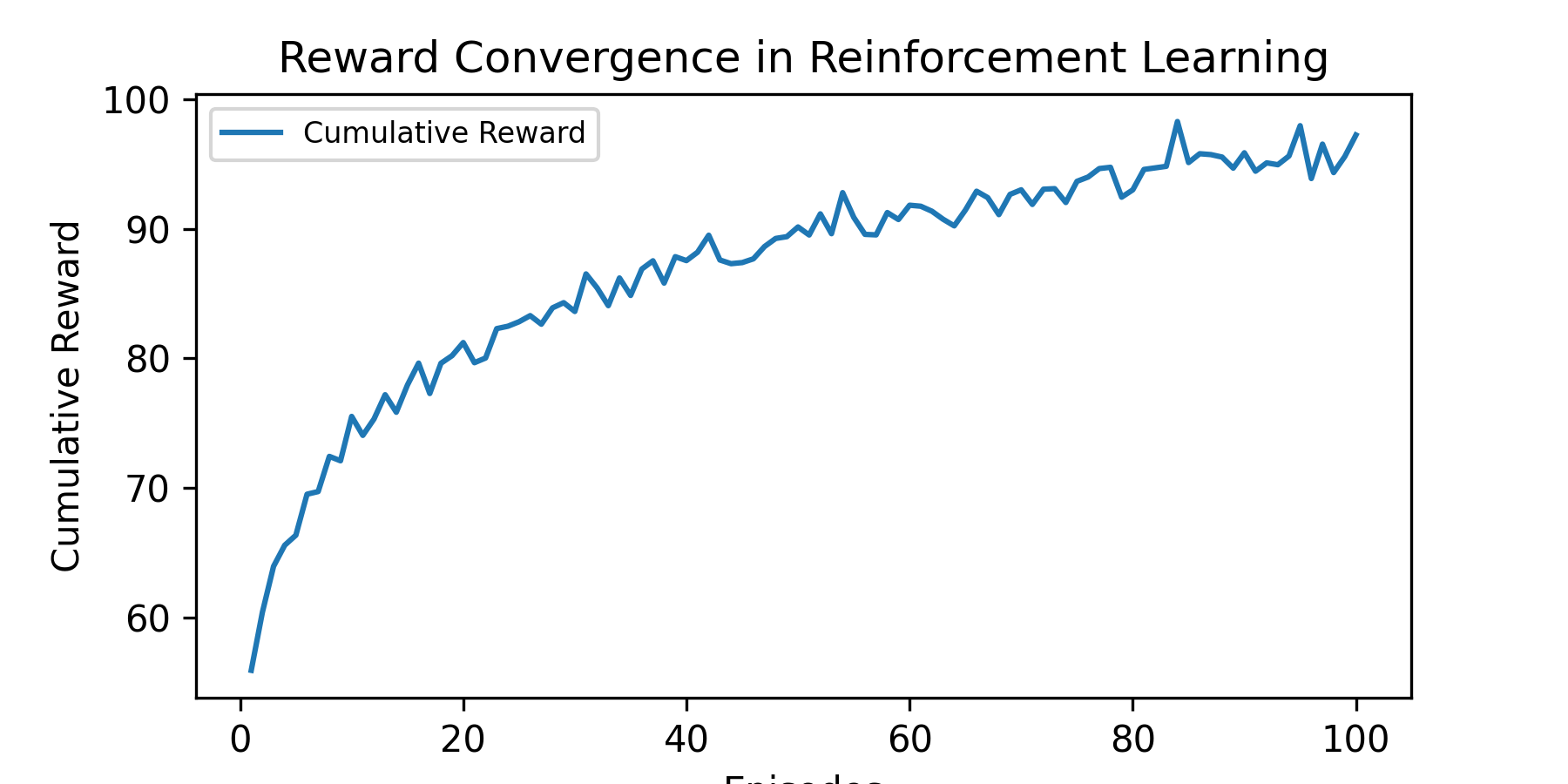}
    \caption{Reward Convergence in Reinforcement Learning}
    \label{fig:reward_convergence}
\end{figure}
As depicted in Figure~\ref{fig:reward_convergence}, the reinforcement learning component effectively learns optimal policies over time. The convergence of rewards demonstrates the system's ability to adapt to diverse vehicular environments and optimize offloading decisions dynamically.

\subsection{Optimization Performance (PSO)}
\begin{figure}[h]
    \centering
    \includegraphics[width=0.45\textwidth, height=0.25\textwidth]{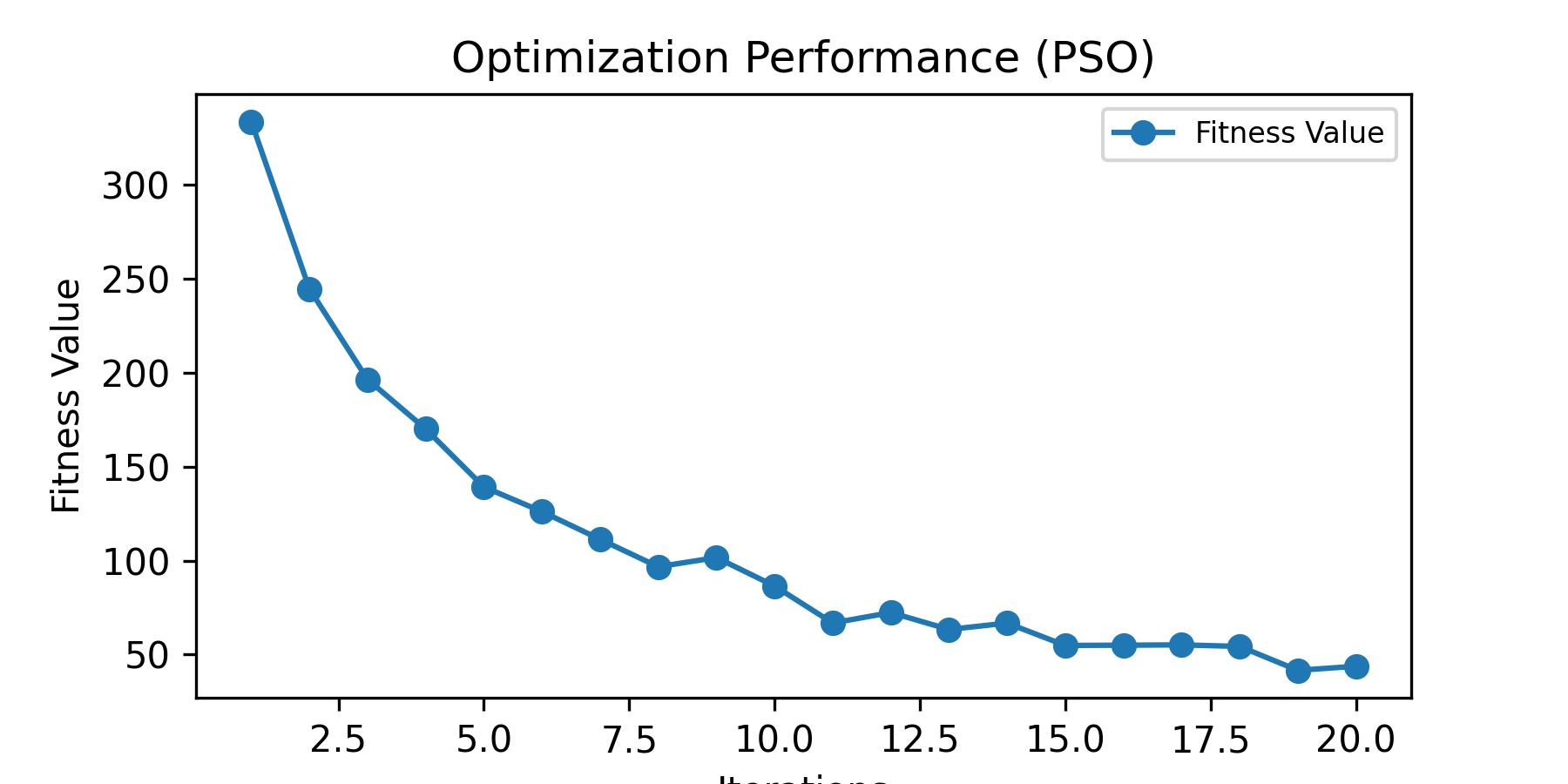}
    \caption{Optimization Performance of PSO}
    \label{fig:optimization_pso}
\end{figure}
Figure~\ref{fig:optimization_pso} illustrates the efficiency of the PSO algorithm in converging to an optimal resource allocation strategy. By balancing latency and energy consumption, the PSO component fine-tunes the overall offloading process, outperforming traditional methods.

\section{conclusion}
This paper proposed a hybrid AI-based task offloading framework for VANETs to address the inherent challenges of dynamic vehicular environments. By integrating supervised learning for predictive analytics, reinforcement learning for adaptive decision-making, and PSO for optimization, the framework effectively reduces latency, minimizes energy consumption, and enhances resource utilization. Simulation results validate the superiority of the proposed approach, demonstrating significant performance gains compared to traditional offloading methods. The framework ability to dynamically adapt to varying network conditions and task demands ensures reliable and efficient offloading even in highly mobile and resource-constrained scenarios. This research contributes to the advancement of VANET technologies, paving the way for their adoption in intelligent transportation systems, autonomous driving, and real-time vehicular applications. Future work will focus on integrating additional AI techniques to address emerging challenges such as cybersecurity and data privacy in vehicular networks.


\begin{thebibliography}{1}
\bibitem{1} J. Yan, S. Bi, and Y. J. A. Zhang, “Offloading and resource allocation
with general task graph in mobile edge computing: A deep reinforcement
learning approach,” IEEE Transactions on Wireless Communications,
vol. 19, no. 8, pp. 5404–5419, 2020.
\bibitem{2} F. Wang, J. Xu, and S. Cui, “Optimal energy allocation and task offload-
ing policy for wireless powered mobile edge computing systems,” IEEE
Transactions on Wireless Communications, vol. 19, no. 4, pp. 2443–
2459, 2020.
\bibitem{3} Z. Zhou, H. Liao, X. Zhao, B. Ai, and M. Guizani, “Reliable task of-
floading for vehicular fog computing under information asymmetry and
information uncertainty,” IEEE Transactions on Vehicular Technology,
vol. 68, no. 9, pp. 8322–8335, 2019.
\bibitem{4} B. Gu, M. Alazab, Z. Lin, X. Zhang, and J. Huang, “Ai-enabled task
offloading for improving quality of computational experience in ultra
dense networks,” ACM Transactions on Internet Technology (TOIT),
vol. 22, no. 3, pp. 1–17, 2022.
\bibitem{5} T. Qayyum, Z. Trabelsi, A. Tariq, M. Ali, K. Hayawi, and I. U. Din,
“Flexible global aggregation and dynamic client selection for federated
learning in internet of vehicles,” Computers, Materials and Continua,
vol. 77, no. 2, p. 1739, 2023.
\bibitem{6} F. Saeik, M. Avgeris, D. Spatharakis, N. Santi, D. Dechouniotis, J. Vio-
los, A. Leivadeas, N. Athanasopoulos, N. Mitton, and S. Papavassiliou,
“Task offloading in edge and cloud computing: A survey on mathe-
matical, artificial intelligence and control theory solutions,” Computer
Networks, vol. 195, p. 108177, 2021.
\bibitem{7} L. Yin, J. Luo, C. Qiu, C. Wang, and Y. Qiao, “Joint task offloading
and resources allocation for hybrid vehicle edge computing systems,”
IEEE Transactions on Intelligent Transportation Systems, vol. 25, no. 8,
pp. 10355–10368, 2024.
\bibitem{8} H. Xiao, C. Xu, Y. Ma, S. Yang, L. Zhong, and G.-M. Muntean, “Edge
intelligence: A computational task offloading scheme for dependent iot
application,” IEEE Transactions on Wireless Communications, vol. 21,
no. 9, pp. 7222–7237, 2022.
\bibitem{9} G. Qu, H. Wu, R. Li, and P. Jiao, “Dmro: A deep meta reinforcement
learning-based task offloading framework for edge-cloud computing,”
IEEE Transactions on Network and Service Management, vol. 18, no. 3,
pp. 3448–3459, 2021.
\bibitem{10} Y. Chen, F. Zhao, X. Chen, and Y. Wu, “Efficient multi-vehicle task of-
floading for mobile edge computing in 6g networks,” IEEE Transactions
on Vehicular Technology, vol. 71, no. 5, pp. 4584–4595, 2022.
\bibitem{11} K. Xiong, S. Leng, C. Huang, C. Yuen, and Y. L. Guan, “Intelligent task
offloading for heterogeneous v2x communications,” IEEE Transactions
on Intelligent Transportation Systems, vol. 22, no. 4, pp. 2226–2238,
2021.
\bibitem{12} A. Tariq, R. A. Rehman, B.-S. Kim, et al., “An intelligent forwarding
strategy in sdn-enabled named-data iov.,” Computers, Materials \&
Continua, vol. 69, no. 3, 2021.
\bibitem{13} Z. Trabelsi, M. Ali, and T. Qayyum, “Fuzzy-based task offloading in
internet of vehicles (iov) edge computing for latency-sensitive applica-
tions,” Internet of Things, vol. 28, p. 101392, 2024.
\bibitem{14} L. Ai, B. Tan, J. Zhang, R. Wang, and J. Wu, “Dynamic offloading
strategy for delay-sensitive task in mobile-edge computing networks,”
IEEE Internet of Things Journal, vol. 10, no. 1, pp. 526–538, 2023.
\bibitem{15} T. Qayyum, A. W. Malik, M. A. Khan, and S. U. Khan, “Modeling
and simulation of distributed fog environment using fognetsim++,” Fog
Computing: Theory and Practice, pp. 293–307, 2020.
\end{thebibliography}

\end{document}